# Deep Residual Networks with a Fully Connected Reconstruction Layer for Single Image Super-Resolution

Yongliang Tang[1], Jiashui Huang[1], Faen Zhang[1*], Weiguo Gong[2]

[1]AInnovation Co. Ltd, Beijing 100089, China

[2]Key Lab of Optoelectronic Technology & Systems of Education Ministry, Chongqing

University, Chongqing 400044, China

Correspondence information: Faen Zhang, AInnovation Co. Ltd, Beijing 100089, China,

zhangfaen@ainnovation.com,

+86 010 57525264

# Deep Residual Networks with a Fully Connected Reconstruction Layer for Single Image Super-Resolution

Yongliang Tang[1], Jiashui Huang[1], Faen Zhang[1*], Weiguo Gong[2]

[1]AInnovation Co. Ltd, Beijing 100089, China

[2]Key Lab of Optoelectronic Technology & Systems of Education Ministry, Chongqing University, Chongqing 400044, China

**Abstract**

Recently, deep neural networks have achieved impressive performance in terms of both reconstruction accuracy and efficiency for single image super-resolution (SISR). However, the network model of these methods is a fully convolutional neural network, which is limit to exploit the differentiated contextual information over the global region of the input image because of the weight sharing in convolution height and width extent. In this paper, we discuss a new SISR architecture where features are extracted in the low-resolution (LR) space, and then we use a fully connected layer which learns an array of differentiated upsampling weights to reconstruct the desired high-resolution (HR) image from the final obtained LR features. By doing so, we effectively exploit the differentiated contextual information over the whole input image region, whilst maintaining the low computational complexity for the overall SR operations. In addition, we introduce an edge difference constraint into our loss function to preserve edges and texture structures. Extensive experiments validate that our SISR method outperforms the existing state-of-the-art methods.

**Keywords:** single image super-resolution, deep neural networks, fully connected reconstruction layer,

Corresponding author. Tel.: +86 010 57525264; E-mail address: zhangfaen@ainnovation.com

edge difference constraint.

# 1 Introduction

Single image super-resolution (SISR), which aims at recovering the visually pleasing high-resolution (HR) image from a single low-resolution (LR) image generated by the low-cost imaging system and the limited environment condition, has gained increasing research attention for decades in computer vision. Since the obtained HR images often preserve important details and critical information for later image processing, analysis and interpretation, SISR is widely applied to various field such as video surveillance [2], medical imaging [3], face recognition [4], satellite imaging [5] and etc.

SISR problem usually assumes the observed LR image to be a non-invertible low-pass filtered, downsampled and noisy version of HR image. Due to the loss of high-frequency information during the degradation of HR images, SISR is a highly ill-posed problem. To handle the ill-posed nature in SR reconstruction, a variety of methods has been developed in computer vision community. Early methods include interpolation and reconstruction-based method. Interpolation methods such as bicubic interpolation [6], edge-guided interpolation [7] and nearest neighbor interpolation [8], usually perform well in smooth areas, while they generate ringing and jagged artifacts in high frequency image regions. Although reconstruction-based methods are effective to preserve sharp edges and suppress ringing artifacts by introducing appropriate image prior knowledge such as edge-directed priors [9], gradient profile priors [10], Bayesian priors [11], and nonlocal self-similarity priors [12], they fails to add sufficient novel high frequency details to the reconstructed HR images with complex scenes.

Currently, learning methods are widely applied to learn the mapping between LR and HR image spaces from millions of co-occurrence LR-HR example image pair, including local linear regression [13], sparse dictionary learning [14], random forest [15], and deep neural network [16, 21]. Among them, the deep neural network (DNN) has drawn considerable attention due to its simple architecture and excellent performance. However, DNN-based SR methods also exhibit limitations in architecture optimality. First, the network model of these methods is a fully convolutional neural network, which is limit to exploit the differentiated contextual information over global image region. Although some methods [18, 19, 22, 23, 24, 33] have improved reconstruction quality by stacking more convolution layers to exploit contextual information over larger image region, they also increases the computation cost and memory usage. Thus, these methods still exhibit limitations in terms of balancing the reconstruction accuracy and efficiency. In addition, these methods use convolution as reconstruction layer to obtain the final HR image, which is limit to utilize the extracted feature information differentially to reconstruct the desired HR images due to the weight sharing of convolution height and width extent. Second, most existing SR algorithms optimize the network models with L2 loss and thus inevitably generate blurry edges and textures in the reconstructed HR images. Several algorithms [22, 23, 24] have focused on improving the loss function to achieve the impressive measures and make the reconstructed HR images close to human visual perception on natural images. However, the blurring problem of sharp edges and texture structures still exist in reconstructed HR images.

To address these drawbacks, we propose a new image SR architecture based on the deep neural networks. Our network takes an LR image as input and trains a cascade of convolutional blocks inspired by deep Residual Networks used for ImageNet classification [25] to extract features in the LR space. Then, we use a fully connected layer which learns an array of upsampling weights to predict residual image (the differences between the upsampled image by bicubic interpolation and the ground truth HR image) from the extracted LR features. Finally, the desired HR image is obtained by adding the predicted residual image to the upsampled image using bicubic interpolation. In addition, considering that L2 loss function used for most SR methods always leads to the blurring of image details and generates ringing artifacts, we introduce an edge difference constraint into the loss function of our proposed network to preserve edges and texture structures.

Overall, the contributions of this paper are mainly in three aspects:

(1) The proposed network extracts directly feature from LR images and jointly optimizes a fully connected upsampling layer to predict residuals image. Since the upsampling layer for our network can utilize all feature information differently to predict each pixel of the HR images, our network is able to differentially exploit context information over the global region of the input image with shallow convolution layers. As a result, our model has a large capacity to learn complicated mappings and effectively reduces the undesired visual artifacts.

(2) Similar to [21], the computation time and memory usage for our networks are sufficiently reduced since our network extracts features directly from LR images and upscales the resolution to HR space in the last layer of the network. However, our networks improve accuracy by exploiting global context information, as illustrated in **Fig.1.** In addition, since all convolution layers can be shared by the networks of the different upscaling factors, our method could facilitate fast training and testing across the different upscaling factors.

(3) We propose a new loss function with an edge difference constraint to optimize the proposed networks for the reconstructed HR images with sharp edges and finer texture details.

## 2 Related work

Numerous methods have been proposed to solve single image SR problem. In this section, we focus our discussion on SR methods that based son deep neural networks.

### 2.1 Deep neural network for SR

In general, the observed LR images are a degraded product of HR images, which can be generally formulated as,

$$y = DHx + v \tag{1}$$

where $x$ and $y$ represent the original HR and observed LR image respectively, $D$ is the downsampling operator, $H$ is the blurring filter, and $v$ represents the additive noise. In view of the above, it is a typical multi-output regression problem to reconstruct a HR image $x$ from the observed LR image. Inspired by the promising performance of deep neural networks in classification and regression tasks, Dong et al. [16] propose a new SR architecture, namely Super-Resolution Convolutional Neural Network (SRCNN). In SRCNN, the mapping $F$ used for reconstructing the desired HR image $x$ consists of three convolution layers and is trained by minimizing the following function,

$$L(\Theta) = \frac{1}{N}\sum_{i=1}^{N}\|F(y_i;\Theta) - x_i\|^2 \tag{2}$$

where $\Theta = \{W_1, W_2, W_3, B_1, B_2, B_3\}$ is the filter and bias of convolution layers in SRCNN, $x_i$ and $y_i$ represent the HR and LR image patch respectively and $N$ is the number of training samples in each batch.

Since the model of SRCNN is shallow network (only including patch extraction/representation, non-linear mapping and reconstruction layer), the prediction of HR image relies on context of small image regions. To exploit more contextual information in a large image region, Kim et al. [18] propose a deep convolutional neural network (CNN) for image SR problem (VDSR) by cascading small convolution layer many times. Although VDSR significantly improves the reconstruction accuracy, the computation time and memory usage also increase with the depth of network. To reduce the computational cost, Dong et al. [21] use a transposed convolution (also named as deconvolution in some of the literature) to upscale the features to HR image space in the last layer of network model. Lai et al. [22] adopt a similar idea and propose a deeper convolutional network within Laplacian pyramid framework (LapSRN) to progressively reconstruct HR images. By doing so, LapSRN improves accuracy without increasing computational burden. However, those kinds of approaches have one limitation: the prediction of each pixel of the desired HR images relies on the context information in a local region of the input images since the model of these method consists of convolutional layers only. In order to exploit context information over the larger region and improve the reconstruction quality, we need to cascade more convolution layers in the networks, which means to increase computational cost and memory usage. In this paper, we discuss a new SR architecture to resolve the dilemma between the reconstruction accuracy and efficiency. The proposed networks not only improve the quality of reconstructed HR images by exploiting the contextual information over the global region of the input image but also reduce the computational cost by using the simplified residual blocks to extract features in the LR space.

**2.2 Loss function**

As in most image restoration tasks, mean squared error (MSE) or L2 loss is also widely used to optimize the network model of image SR. Since L2 loss is the major performance measure (PNSR) for those problem, the trained models usually have impressive performance in terms of objective measure.

However, there is a blurring problem of texture details and edge structures in reconstructed HR images. Several studies have focused on the loss functions to better train network models and restore finer texture details and sharp edges. Inspired by the report that training with L2 loss cannot guarantee better performance compared to other loss functions in terms of PSNR and SSIM [26], Lim et al. [24] use L1 loss to optimize their network models for achieving improved performance. Lai et al. [22] propose a robust Charbonnier loss for the deep convolutional network within Laplacian pyramid framework (LapSRN). At each pyramid level, LapSRN has corresponding loss function to reduce the difference between output reconstructed images and the label image $x_s$ downsampled from ground truth HR image with bicubic interpolation. Accordingly, the overall loss of LapSRN is defined as

$$L(\Theta) = \frac{1}{N}\sum_{i=1}^{N}\sum_{s}^{L} \rho\left(\left(F(y^i;\Theta) + y_s^i\right) - x_s^i\right) \qquad (3)$$

where $\rho(\cdot)$ is the Charbonnier penalty function, $L$ is the number of pyramid level, and $y_s$ is the upsampled image from the input LR image $y$ in the pyramid level $s$. Due to the deep supervision of multi-loss structure and the robustness of Charbonnier penalty function, Charbonnier loss improves the stability of networks training and the reconstruction quality. Ledig et al. [23] propose a perceptual loss function which consists of an adversarial loss and a content loss to reconstruct plausible-looking natural HR images with high perceptual quality.

$$L(\Theta) = \frac{1}{N}\sum_{i=1}^{N}\left\|\phi\left(G_\theta(y_i)\right) - \phi(x_i)\right\|^2 - \log D_\theta\left(G_\theta(y_i)\right) + \left\|\nabla G_\theta(y_i)\right\| \qquad (4)$$

where $\phi(\cdot)$ is the feature representations, $D_\theta\left(G_\theta(y_i)\right)$ is the estimated probability that the reconstructed HR image is a natural image and $\nabla G_\theta(y_i)$ is a regularizer based on the total variation to encourage spatially coherent solution. Although these researches of loss function have improved the quality of reconstructed HR images, the restoration of finer texture details and sharp edges is still a challenging problem. Accordingly, we propose a new loss function with the edge difference constraint to reconstruct edges and texture details

## 3 Proposed method

In this section, we detail the methodology of the proposed deep neural network, the loss functions with the edge difference constraint, and the training of our network.

### 3.1 Network architecture

As shown in Fig.2, our network can be decomposed into two parts – features extraction and SR reconstruction. The part of features extraction takes an observed LR image $y$ as input and uses the cascaded residual building blocks to extract features in LR space. SR reconstruction part is a fully connected layer, which upsamples and aggregates the previous features with an array of trainable weights to reconstruct the desired HR images. In the following sections, we first describe the residual units of our networks, and then we suggest the single upscaling model that handles a specific SR scale and the multi-upscaling strategy that quickly trains models for reconstructing various upscaling of HR images.

**Residual units.** Deep residual networks [1] have emerged as a family of extremely deep architectures showing compelling accuracy and nice convergence behaviors in computer vision, machine translation, speech synthesis, speech recognition. Although the deep residual architecture has been successfully applied to the image SR problem and exhibited excellent performance [23, 24], we further improve the performance by employing a new residual unit which makes training easiness and reduces training error.

In Fig. 3, we show the residual units of each network from SRResNet [23], EDSR [24], and ours. Although all the skip connections and after-addition activation functions for the residual units of these networks are the identity mapping for creating a direct path to propagate information - not only within a residual unit, but through the entire network, only our residual units adopt a re-arranging the after-addition activation method to directly propagate information from one unit to any other units in both forward and backward passes. SRResNet removes the Rectified Liner Unit (ReLU) to make after-addition activation into an identity mapping. EDSR improves SRResNet by removing the Batch Normalization (BN) layers to enhance the flexibility of the network and reduce GPU memory usage. Inspired by [25], our networks recast the after-addition activation as the pre-activation of the next residual unit, which means that the activation only affects the residual function (Fig. 3(c)). By re-arranging the after-addition activation, we not only reduce the difficulty of network optimization because of the identity mapping, but improves regularization of the models since we don't remove ReLU layers from our networks. At the same time, we remove BN layers from our networks since they consume the same amount of memory as the preceding convolution layers.

Furthermore, we use the Parametric Rectified Liner Unit (PReLU) to instead of ReLU as the activation function of our convolution layer. Since PReLU has a learnable coefficient for the negative part of features, it can void the "dead features" cause by zero gradients in ReLU. Accordingly, we can make full use of all parameters to obtain the maximum capacity of our networks.

**Single upscaling model.** In the convolutional networks, model performance can be enhanced by cascading multiple small filters in a deep network structure. Thus, we further improve our residual unit (Fig. 3(d)) with bottleneck [25] architecture and use it to construct the feature extraction part of our single upscaling model. A bottleneck residual unit consists of a $1 \times 1$ layer for reducing dimension, a $3 \times 3$ layer, and a $1 \times 1$ layer for restoring dimension. As designed in [25], its computational complexity is similar to the residual unit including two $3 \times 3$ convolution layers (Fig. 3(c)). However, the model with the bottleneck residual units has improved performance due to the increase of network depth. Therefore, we can maximize our model capacity considering the limited computational resources.

For original residual networks, the convolution shortcuts [25] are used to reduce the feature map size and increase dimensions. Since the convolution shortcut is not an identity mapping, the direct path for propagating information is limited to a local region with the same feature map size. Although our network can create a direct path for propagating information over the entire network since it takes LR images as input and extracts all the features in LR space, we found that stacking the number of residual units above a certain level would make the training procedure numerically unstable. We resolve this issue by incorporating the convolution layers into the cascaded residual units and constructing local paths for propagating information directly. As shown in Fig.2, we use three residual units and one convolution layer to form one residual block which has a local propagating path. To void the "dead features" in the identity path, we remove the activation (ReLU) of the incorporated convolution layers from residual blocks, as illustrated in Fig.2.

Since the existing HR reconstruction layers (the transposed convolution [21] or sub-pixel convolution [24] ) only use the feature information in a very small local region to predict each pixel in reconstructed HR images and also share the weights when predicting all the pixels of the reconstructed HR images, the reconstruction results usually contain undesired artifacts. To make effective use of the extracted features and improve reconstruction accuracy, we propose a fully connected reconstruction layer to differentially utilize the extracted feature information over the global region. However, the parameters of our reconstruction layer would be significantly increased since the fully connected layer construct the trainable weights between each predicted HR pixel and all the extracted features. We resolve the issue by reducing the dimension of the extracted feature map by using the incorporated convolution layer in the final residual blocks.

Our final network models are constructed following. The number of residual blocks and the dimension of feature maps in the identity path are set to be 5 and 128, respectively. In residual units, we reduce the dimension to 64 for bottleneck architecture. Our final dimension of the extracted feature maps will shrink to 8 for reducing the total parameters of our networks and improving the training and testing efficiency.

**Different upscaling factors.** In reality SR applications, we usually need to reconstruct various upscaling factors of HR images. Thus, we expect the proposed method could achieve fast training and testing across different upscaling factors. Since all convolution layers on the whole act like a complex feature extractor of the LR image, and only the last reconstruction layer contains the information of the upscaling factor, we can transfer the convolution filters for fast training and testing.

In practice, we train a model for an upscaling factor in advance. Then, we only fine-tune the reconstruction layer for another upscaling factor and leave the convolution layers unchanged. The fine-tuning is fast, and the performance is as good as training from scratch (see Section 4.1). During testing, we perform the convolution operations once, and upsample an image to different sizes with the corresponding reconstruction layer. Furthermore, our method can reconstruct HR images with an arbitrary resolution (non-integer upscaling factor) by fine-tuning the fully connected reconstruction layer.

**3.2 Loss function**

For most of DNN-based SR methods, L2 is the most widely used loss function. Thus, the optimization objective for these networks is to minimize the following function,

$$\min_{\Theta} \frac{1}{N} \sum_{i=1}^{N} \|F(\boldsymbol{y}^i; \Theta) - \boldsymbol{x}^i\|_2^2 \tag{5}$$

where $\boldsymbol{y}^i$ and $\boldsymbol{x}^i$ are $i$-th LR and HR image pair in the training data, and $F(\boldsymbol{y}^i; \Theta)$ is the predicted HR image using the training network with parameters $\Theta$. Since L2 loss struggles to handle the uncertainty inherent relationship in recovering lost high-frequency details such as small scale structures and texture details, it encourages finding pixel-wise averages of plausible solutions which are typically overly-smooth and thus have poor perceptual quality [27]. In order to resolve this problem, we propose a new loss function for our network. Similar to [28], we proposed loss use the following edge difference constraint to preserve edges and texture structures,

$$\boldsymbol{E}_d = \|\mathrm{E}(\boldsymbol{x}_h) - \mathrm{E}(\boldsymbol{x})\|_p \tag{6}$$

where $E(\cdot)$ is the edges and textures extraction operation, $\boldsymbol{x}_h$ is reconstructed HR images, and $p$ is the norm of the edge difference constraint. Considering that the one-dimensional (1-D) processing of images can provide effective edge and texture information and handle outliers [28], we use 1-D processing as the edges and textures extraction operator. Actually, for a given image $\boldsymbol{x}$, the extraction of edges and textures information can be formulated as,

$$E(\boldsymbol{x}) = \sqrt{H(\boldsymbol{x})^2 + V(\boldsymbol{x})^2} \tag{7}$$

where $H(\boldsymbol{x})$ and $V(\boldsymbol{x})$ are the horizontal and vertical edge and texture information of the image $\boldsymbol{x}$ respectively. 1-D first use the Gaussian operator to smooth horizon direction, and then the first derivative of Gaussian operator is applied along the orthogonal direction for obtaining the vertical edges and textures $V(\boldsymbol{x})$. By repeating this procedure, we can obtain the corresponding horizon edge and texture information $H(x)$. Since the smoothing is done along a direction orthogonal to the direction of the edge extraction, 1-D processing can effectively handle outliers and extract image edges and textures. Thus, the trained network with edge difference constraint can generate sharp edges and finer texture details in the reconstructed HR images. By incorporating the edge difference constraint into L2 loss, we can obtain the following optimization objective for our network models,

$$\min_{\Theta} \frac{1}{N} \sum_{i=1}^{N} \|F(y^i; \Theta) - x^i\|_2^2 + \beta \left\| E\left(F(y^i; \Theta)\right) - E(x^i) \right\|_p \tag{8}$$

where $\beta$ is the weight for the edge difference constraint. Since our network also adopt the residual learning, the overall loss function is defined as,

$$L(\Theta) = \sum_{i=1}^{N} \|(F(y^i; \Theta) + x_b^i) - x^i\|_2^2 + \beta \left\| E\left((F(y^i; \Theta) + x_b^i)\right) - E(x^i) \right\|_p \tag{9}$$

here $x_b^i$ is upsampled image from the input LR images $y^i$ using bicubic interpolation. Considering that the researchers [22, 24] report that training with L1 norm achieve improved performance compared with the training with L2, we empirically set $p$ to 1 for our edge difference constraint.

**3.3 Training details**

**Training dataset:** For fair comparison with most state-of-the-art methods, we first use 91 images from Yang et al.[14] and 200 images from the training set of BSD500 [18] as the original images to train our SR models. In addition, considering that big data can push a deep model to the best performance, we also use images from DIV2K [34] datasets to optimize our models and compare with the other state-of-the-art SR models trained with the same dataset. We adopt the following ways to augment the training images: (1) Scaling: each HR image is downsampled by bicubic interpolation with the scaling factor 0.9, 0.8, 0.7, and 0.6. (2) Rotation: each image is rotated with the degree of 90, 180 and 270. (3) Flipping: each image is flipped with horizontal and vertical. Thus, we obtain $5 \times 4 \times 3 = 60$ times image to form the final ground-truth HR image training set $\{X\}$. In order to prepare the training data, we first downsample the HR training images $\{X\}$ with the desired upscaling factor $n$ to form the corresponding LR image $\{Y\}$. Then, we crop the LR training image into a set of LR image patches $\{y^i\}_{i=1}^{N}$ with a stride $k$. The corresponding HR image patches $\{x^i\}_{i=1}^{N}$ are also cropped with a stride $n \times k$ from the HR images. Actually, the cropped LR/HR image patch pairs $\{(y^i, x^i)\}_{i=1}^{N}$ are the training data for the proposed network. Since all the convolutional layers can be shared by the networks of different upscaling factors, it is necessary to employ the LR image patches with the same size to all the networks of our method. Thus, for $\times 2$, $\times 3$ and $\times 4$ networks, the size of LR/HR image patches are set to be $32^2/64^2$, $32^2/96^2$ and $32^2/128^2$, respectively.

**Training Parameters:** For the proposed SR method, we use Caffe package [30] with stochastic gradient descent algorithm to train our networks. For training model from scratch, we use a learning rate of 0.1 for the convolution layers and 0.01 for the fully connected reconstruction layer. The learning rate will be decayed every 10 epochs using a factor of 10. Since we adopt an extremely high learning rates (0.1) to accelerate the convergence, the gradient clipping is set to be 1 and then is decreased by a factor of 10 every 10 epochs. For weights initialization, all the filters of the convolution and the fully connected reconstruction layer are initialized with the method described in [31]. During the fine-tuning of another upscaling networks, the learning rate for all layers is set to be 0.001 and decayed by an exponential rate of 0.90 each epoch. The training of scratch network uses momentum with a decay of 0.9, while our fine-tuned models are achieved using RMSProp with decay of 0.9 and $\epsilon = 1.0$. The batches of size and weight decay are set to 256 and 0.0001 for all the network training, respectively

# 4 Experimental results and discussions

In this section, we first analyze the contributions of the different components of the proposed network. Then, we compare the proposed method with state-of-the-art SR algorithms on the representative image datasets. Finally, we discuss the applications of our models on the real-world images.

In our experiments, we follow the publicly available evaluation framework of Timofte et al.[13]. It enables the comparison of the proposed method with many state-of-the-art SR methods in the same setting. The framework only applies the SR reconstruction algorithm on the luminance channel and directly upscales the chrominance (Cb and Cr) channels to the desired resolution using bicubic interpolation. Furthermore, SR performance metrics including the peak signal-to-noise ratio (PSNR) and structural similarity (SSIM) are adopted to evaluate the objective quality of reconstructed HR images.

## 4.1 Investigation of the network model

In this section, we perform experiments to analyze the property of the proposed network and confirm the contributions of the different components of our networks for the accuracy of SR reconstruction.

**Fully connected reconstruction layer.** To demonstrate the effect of our reconstruction layer, we remove the fully connected layer and use the transposed convolution as the reconstruction layer of our networks. For a fair comparison, the proposed network and transposed convolutional network are both optimized using L2 loss function from the same scratch. Fig. 4 shows the convergence curves in terms of PSNR on the Set14 for the upscaling factor of 3. The performance of the transposed convolutional network (purple curve) is significantly worse than the network with a fully connected layer. In view of the above, the proposed network with a fully connected layer is more capable of reconstructing HR images.

**Residual units.** For our proposed models, we use a new residual unit to improve the SR accuracy. Therefore, we verify the effectiveness of our residual unit in the section. To this end, we remove our residual units and use the residual units of EDSR [24] to construct the SR model. For a fair comparison and save time, we also use L2 loss function to optimize the SR model from the scratch initialized with the same method. As illustrated in Fig. 4, the SR model constructed with the residual of EDSR [24] (coffee curve) fluctuates obviously and converges to a worse result slowly. The SR model with our residual units (yellow curve) has better convergence stability and accuracy. This is because our residual units use the re-arranging of the after-addition activation to reduce the difficulty of networks optimization and improve the regularization of our SR models.

**Loss function.** In our method, we present a new loss function to preserve edges and texture structures. Here we verify the effectiveness of the proposed loss function. For comparison, we use L2 loss function to optimize our network in the same training parameters and initialization method. As illustrated in Fig. 4, the network optimized with L2 loss (yellow curve) converges smoothly, but has high training loss. Although our final model (red curve) fluctuates "significantly", it can obtain the improved performance in terms of reconstruction accuracy. This is mainly because that our loss applies more effective constraints among the predicted images and ground-truth HR training images.

**Different upscaling factors.** In this section, we demonstrate the flexibility of our network for fast training and testing across different upscaling factors. In our experiment, we first obtain a well-trained model with the upscaling factor of 3, then we train the network for $\times 2$ on the basis of that for $\times 3$. During training, we only fine-tune the fully connected reconstruction layer on the training datasets of

×2 since the parameters of all convolution filters in the well-trained model are transferred to the new network. For comparison, we also train another network for ×2 but from scratch. The convergence curves of these two networks are shown in Fig. 5. Obviously, with the transferred parameters, the network converges very fast (only a few epochs) with the same good performance as that training from scratch. In the following experiments, we only train the networks from scratch for ×3, and fine-tune the corresponding reconstruction layers for ×2 and ×4.

**4.2 Comparisons with the state-of-the-arts methods**

To validate the performance of the proposed method, the image SR experiments of different scaling factors (×2, ×3 and ×4) are performed on all the images in the five representative image datasets Set5, Set14, BSD100, Urban100 and Manga109 [22]. Among these datasets, Set5, Set14 and BSD100 consist of natural scenes images; Manga109 and Urban100 include challenging images with details in different frequency bands. Then, we compare the proposed method with other state-of-the-art SR algorithms: FSRCNN [21], VDSR [18], DRCN [19], DRRN [33], LapSRN [22], MemNet [36], EDSR[24], D-DBPN [37] and SRFBN [38]. All the compared results are reproduced by the corresponding public codes under the same setting with our experiments.

We first train our SR models using the images for 291-images dataset and compare with the state-of-the-art SR methods (FSRCNN [21], VDSR [18], DRRN [33] and LapSRN [22]) that also are trained by 291-images. Table 1 shows the average PSNR and SSIM results of reconstructed HR images on the five representative image datasets for the different scaling factors(×2, ×3 and ×4). From Table 1, we can see that the proposed method achieves the consistent performance on all the datasets. Due to the limitations of the training images, the performance of all methods tends to decline with the increase of the test images. However, the proposed method still performs better than all the compared methods in both PSNR and SSIM. These experimental results indicate that the proposed method can effectively improve the quality of reconstructed HR images.

Furthermore, we use more images from DIV2K [34] datasets to push our SR models to the best performance. The final quantitative results are shown in Table 2. From the Table 2, our SR models achieve an improved performance with the increase of training images and have comparable PSNR and SSIM results with the comparative SR methods: MemNet [36], EDSR [24], D-DBPN [37] and SRFBN [38]. Compared with our SR method, EDSR methods uses much more filters (256 vs. 128) in each convolution layer to construct SR models, and D-DBPN and SRFBN employ more training images (DIV2K [34] + Flickr2K [24] + ImageNet [36] and DIV2K [34] + Flickr2K [24] vs. DIV2K [34]) to optimize their SR models. However, our SR method still can earn competitive results with these state-of-the-art SR methods.

In Fig. 6-8, we show visual comparisons on the images, drawn from B100, Urban100 and Manga109, with the upscaling factors of ×4. As shown in Fig. 6-8, our SR model can reconstruct the desired HR images more accurately. For the '108005' image from B100 dataset, FSRCNN, VDSR and DRCN fail to recover the clear stripes on tiger in the reconstructed HR images. Although DRRN and LapSRN provide more clear stripes, the results are much smoother. Our SR models can produce clear and sharp SR images which are very close to the ground-truth HR images by using the proposed residual units, reconstruction layer and loss function.

**4.3 Super-resolving on real-world images**

In this section, we further validate the super-resolving performance of the proposed method on the historical photographs with JPEG compression artifacts. Because neither the ground-truth HR images nor the downsampling kernels are available, our experiment can demonstrate the super-resolving performance of the proposed and compared methods on the real-word images. Fig.9-10 show the super-resolved historical images of the upscaling factor ×4. As shown in Fig.9-10, the proposed method can provide clearer details and sharper edges in the reconstructed HR images than the compared methods.

## 5 Conclusions

In this work, we discuss a new SR architecture based on deep neural networks to reconstruct the desired HR images. By cascading the improved residual blocks to extract features in LR space and jointly optimizing a fully connected reconstruction layer to exploit the differentiated contextual information over the global region of the input LR images, the proposed networks can alleviate the issues of current SR network models and obtain the visually pleasing HR images with the low computational cost. In addition, we propose a new loss function with the edge difference constraint to preserve sharp edge and restore finer texture structures. Experimental results on the images, drawn from five representative image datasets, demonstrate that the proposed networks perform competitively with the existing methods in terms of balancing the reconstruction accuracy and efficiency.

## Acknowledgements

This work was supported by Key Projects of the National Science and Technology Program, China (Grant No. 2013GS500303), the Key Science and Technology Projects of CSTC, China (Grant Nos. CSTC2012GG-YYJSB40001, CSTC2013-JCSF40009).

The authors would like to thank the editors and reviewers for their valuable comments and suggestions.

# Author Contributions

This manuscript was performed in collaboration between the authors. Faen Zhang is the corresponding author of this research work. Yongliang proposed the new SISR method based on deep neural networks. Jiashui Huang and Weiguo Gong were involved in the writing and argumentation of the manuscript. All authors discussed and approved the final manuscript.

Y. Tang et al.[30] Jia, Y., Shelhamer, E., Donahue, et al. T.: Caffe: Convolutional architecture for fast feature embedding. In: ACM MM. (2014) 675–678

[31] K. He, X. Zhang, S. Ren, and J. Sun. Delving deep intorectifiers: Surpassing human-level performance on ImageNet classification. CoRR, abs/1502.01852, 2015.

[32] J.-B. Huang, A. Singh, and N. Ahuja. Single image super-resolution from transformed self-exemplars. In CVPR, 2015.

[33] Y. Tai, J. Yang, and X. Liu. Image super-resolution via deep recursive residual network. In CVPR, 2017.

[34] R. Timofte, E. Agustsson, L. Van Gool, M.-H. Yang, L. Zhang, et al. Ntire 2017 challenge on single image super resolution: Methods and results. In CVPR 2017 Workshops.

[35] O. Russakovsky, J. Deng, H. Su, J. Krause, S. Satheesh, S. Ma, Z. Huang, A. Karpathy, A. Khosla, M. Bernstein, et al. ImageNet large scale visual recognition challenge. International Journal of Computer Vision, 115(3):211–252, 2015.

[36] T. Tong, G. Li, X. Liu, and Q. Gao. Image super-resolution using dense skip connections. In ICCV, 2017.

[37] M. Haris, G. Shakhnarovich, and N. Ukita. Deep backprojection networks for super-resolution. In CVPR, 2018.

[38] Zhen Li, Jinglei Yang, Zheng Liu, Xiaomin Yang and Gwanggil Jeon. Feedback Network for Image Super-Resolution. In CVPR, 2018
23

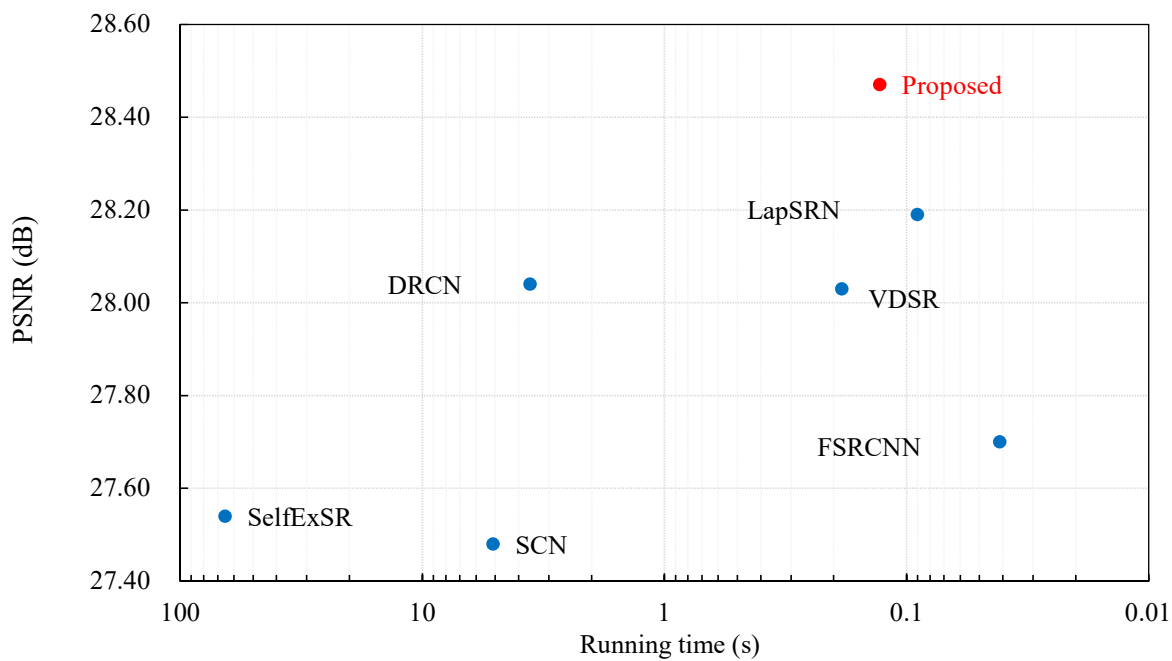

**Fig.1** Average execution time versus PSNR between the proposed and existing methods. The results are evaluated on all the images in Set14 with the upscaling factor of 4. Our method provides the best reconstruction results and preserves the execution time of SelfExSR, SCN, DRCN, VDSR, and LapSRN. All models are trained on the 291-image dataset.

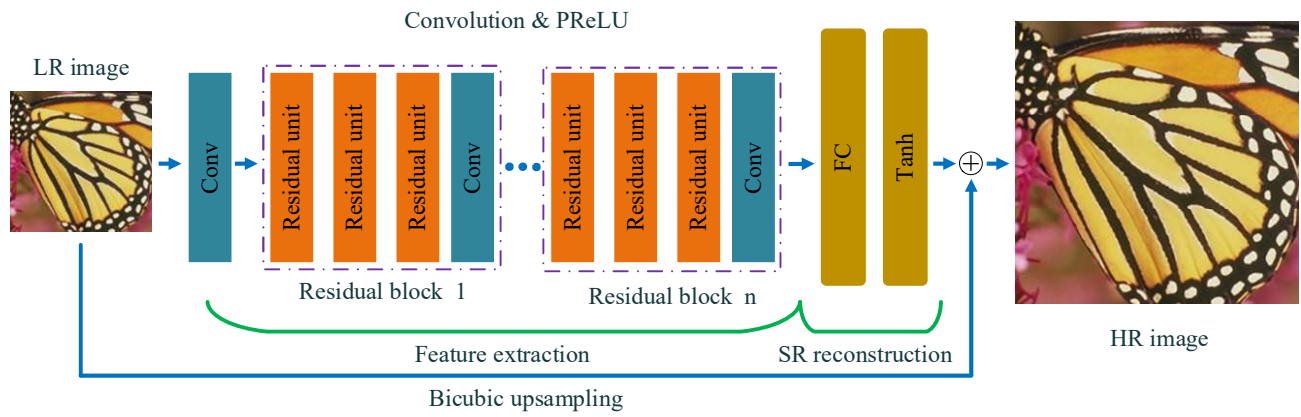

**Fig. 2** The architecture of our proposed SISR network. Our network consists of feature extraction and SR reconstruction.

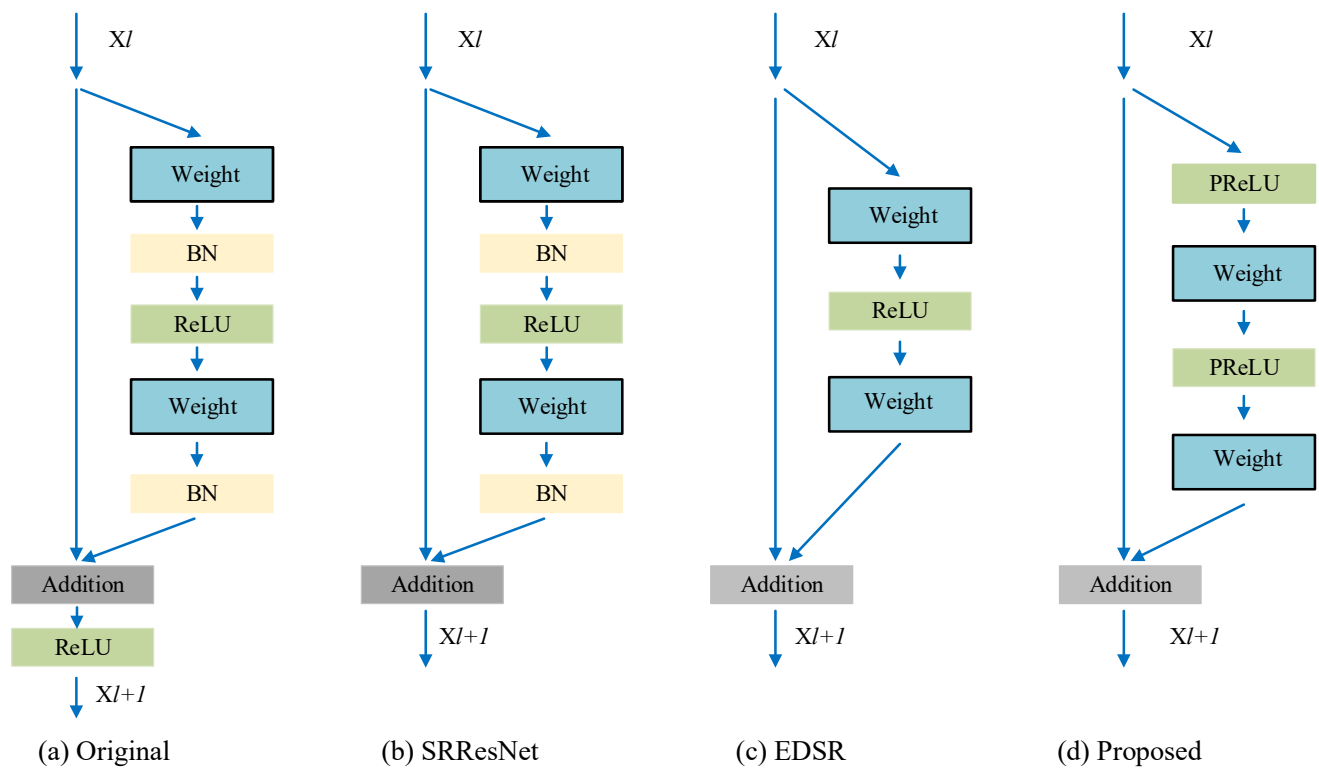

**Fig. 3** Comparison of residual units in Original, SRResNet, EDSR, and ours.

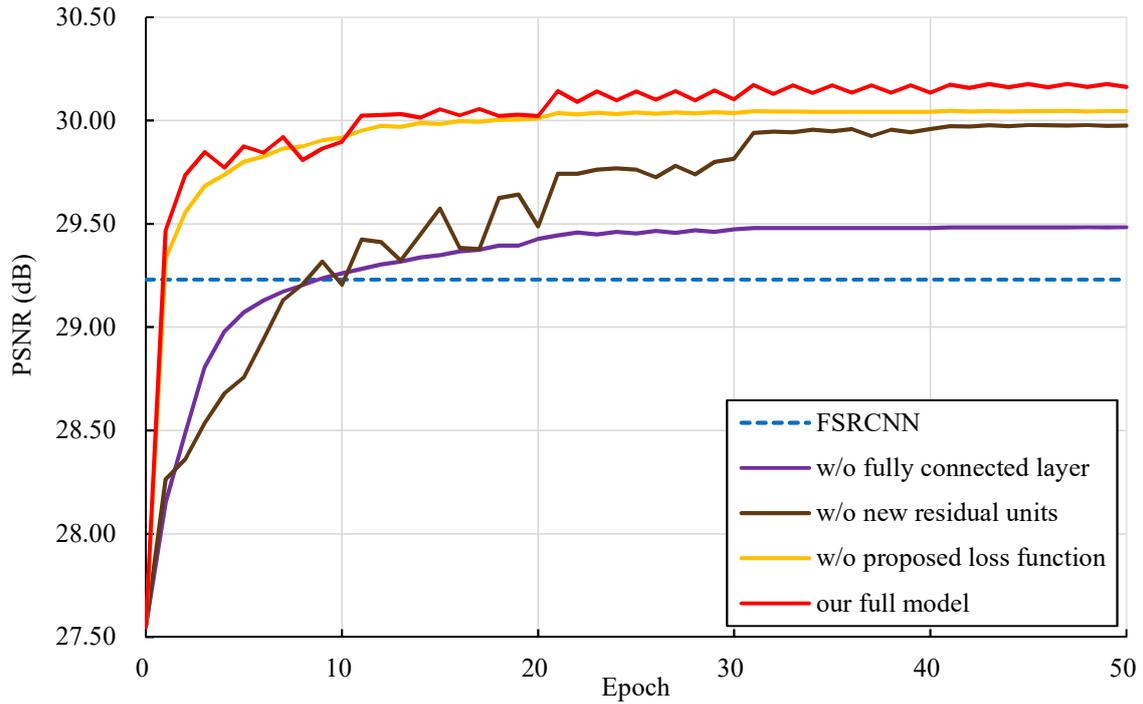

**Fig. 4** Convergence analysis on the different components of the proposed network. The results are obtained on all the images in Set14 with the upscaling factor of ×3. All models are trained on the 291-image dataset.

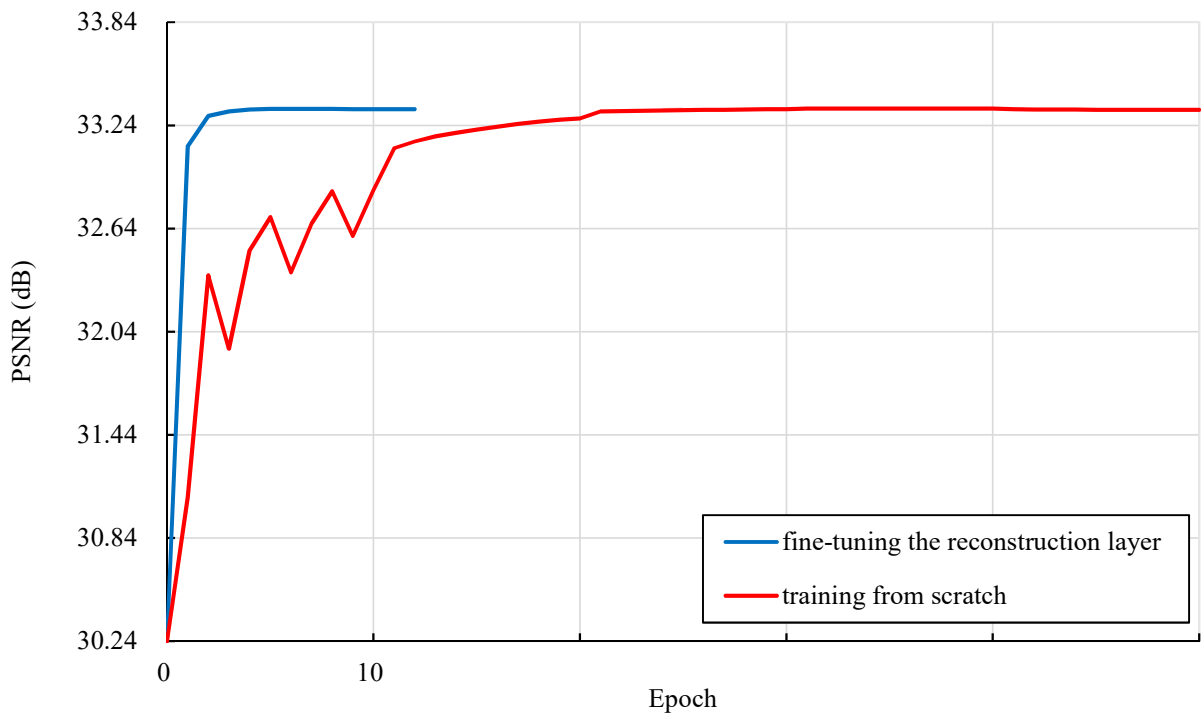

**Fig. 5** Convergence curves of different training strategies. The results are obtained on Set14 with the upscaling factor of ×2. All models are trained on the 291-image dataset.

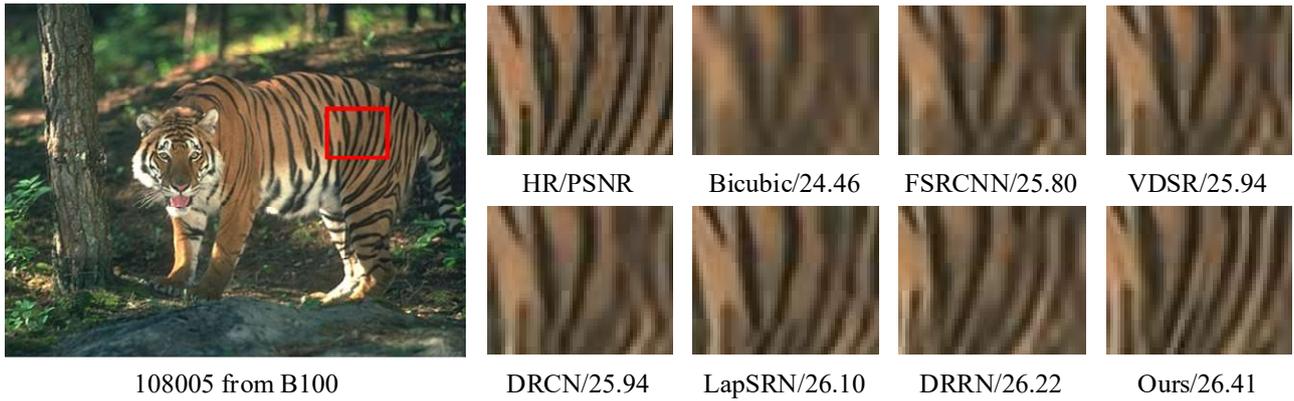

**Fig.6** Visual comparison of our method and the compared methods on image "108005" for the upscaling factor of × 4. All models are trained on the 291-image dataset.

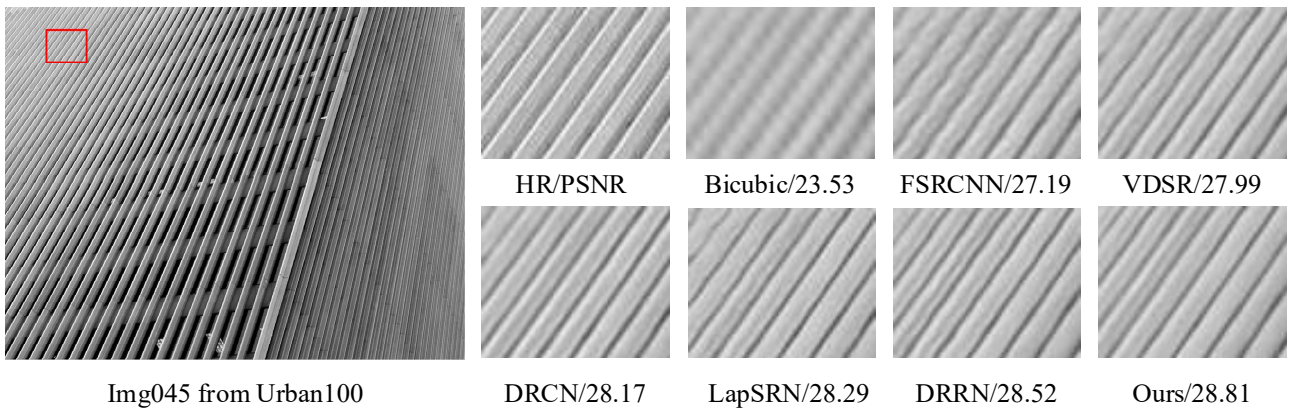

**Fig.7** Visual comparison of our method and the compared methods on image "Img045" for the upscaling factor of × 4. All models are trained on the 291-image dataset.

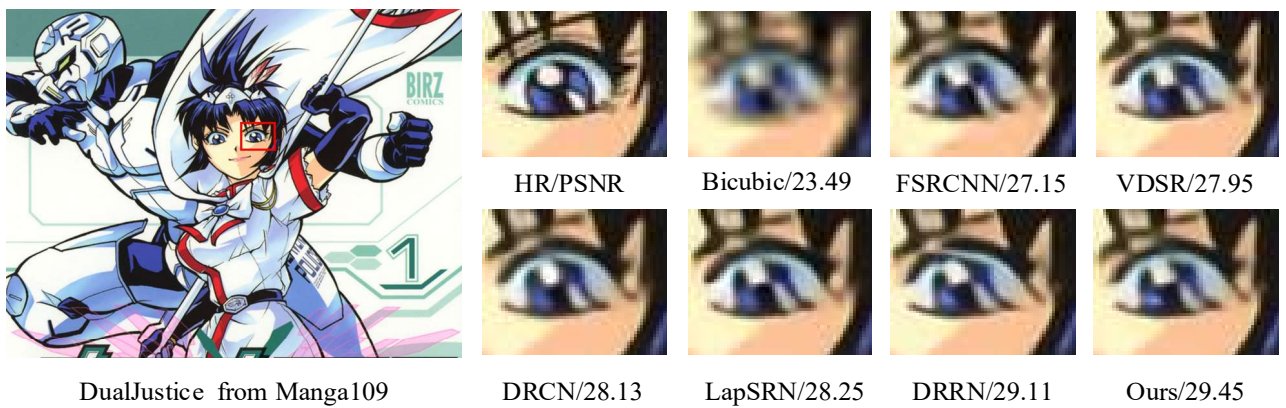

**Fig.8** Visual comparison of our method and the compared methods on image "DualJustice" for the upscaling factor

of ×4. All models are trained on the 291-image dataset.

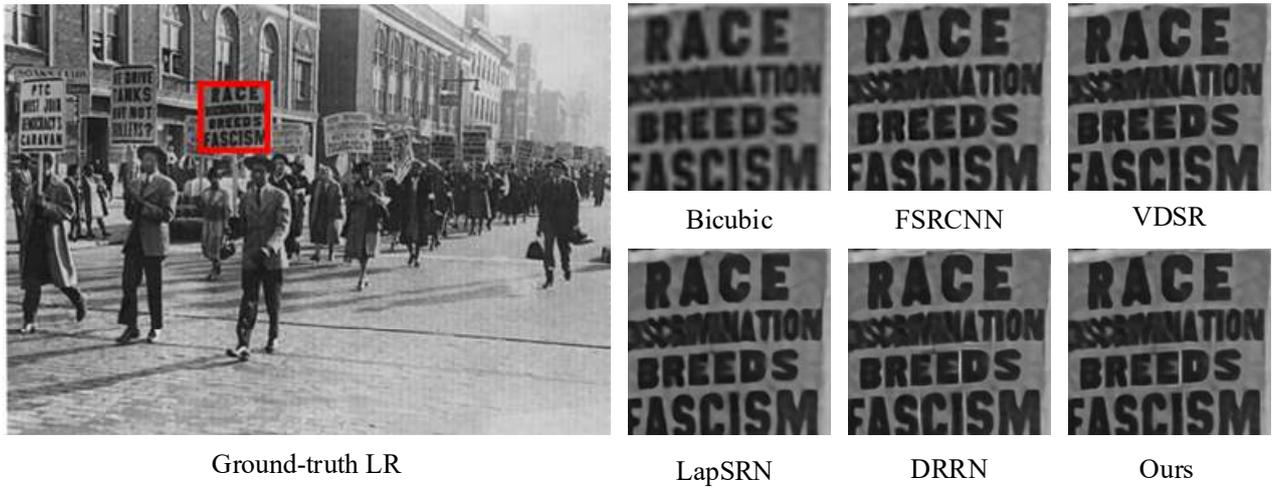

**Fig.9** Visual comparison for ×4 upscaling factor on real-world historical images. All models are SR trained on the 291-image dataset. All models are trained on the 291-image dataset.

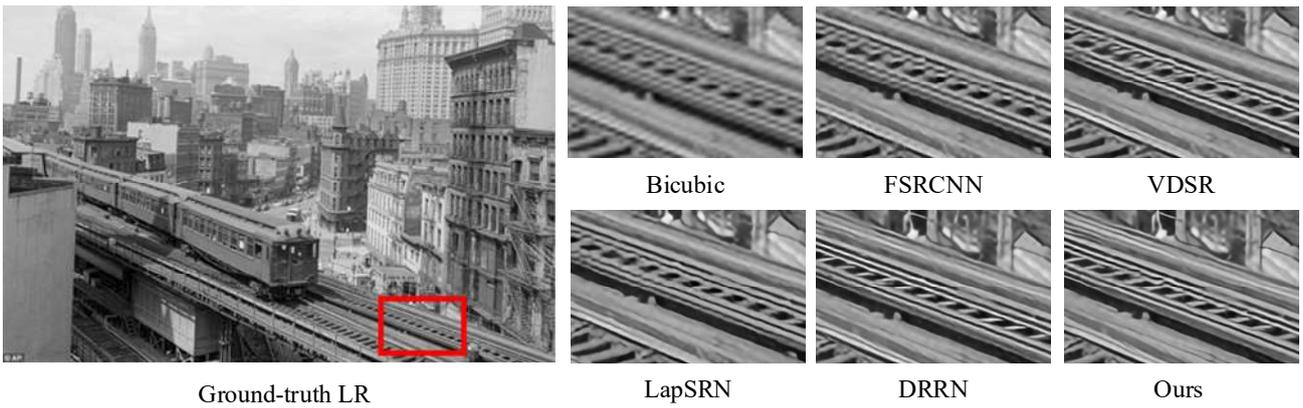

**Fig.10** Visual comparison for ×4 upscaling factor on real-world historical images. All models are trained on the 291-image dataset. All models are trained on the 291-image dataset.

**Table 1** Average PSNR/SSIM for the upscaling factors ×2, ×3 and ×4 on datasets Set5, Set14, B100, Urban100 and Manga109. Red color indicates the best performance and blue color indicates the second-best performance. All models are trained on the 291-image dataset.

| Dataset | Scale | Bicubic | FSRCNN | VDSR | LapSRN | DRRN | Proposed |
|---|---|---|---|---|---|---|---|
| Set5 | x2 | 33.66/0.9299 | 37.00/0.9557 | 37.53/0.9587 | 37.52/0.9591 | 37.74/0.9591 | 37.89/0.9602 |
| | x3 | 30.39/0.8677 | 33.16/0.9132 | 33.66/0.9213 | 33.82/0.9227 | 34.03/0.9244 | 34.22/0.9267 |
| | x4 | 28.42/0.8099 | 30.71/0.8647 | 31.35/0.8838 | 31.54/0.8854 | 31.68/0.8888 | 31.93/0.8917 |
| Set14 | x2 | 30.23/0.8689 | 32.63/0.9087 | 32.97/0.9127 | 33.08/0.9130 | 33.23/0.9136 | 33.39/0.9149 |
| | x3 | 27.54/0.7742 | 29.43/0.8245 | 29.77/0.8314 | 29.79/0.8320 | 29.96/0.8349 | 30.18/0.8382 |
| | x4 | 26.00/0.7026 | 27.59/0.7540 | 28.03/0.7678 | 28.19/0.7720 | 28.21/0.7721 | 28.47/0.7799 |
| B100 | x2 | 29.55/0.8438 | 31.50/0.8909 | 31.89/0.8958 | 31.78/0.8941 | 32.05/0.8973 | 32.17/0.8996 |
| | x3 | 27.21/0.7397 | 28.52/0.7900 | 28.82/0.7976 | 28.82/0.7973 | 28.95/0.8004 | 29.09/0.8083 |
| | x4 | 25.96/0.6693 | 26.97/0.7140 | 27.29/0.7252 | 27.32/0.7280 | 27.38/0.7284 | 27.50/0.7317 |
| Urban 100 | x2 | 26.88/0.8406 | 29.85/0.9011 | 30.77/0.9141 | 30.41/0.9093 | 31.23/0.9188 | 31.53/0.9236 |
| | x3 | 24.46/0.7354 | 26.42/0.8070 | 27.14/0.8279 | 27.07/0.8272 | 27.53/0.8378 | 27.82/0.8422 |
| | x4 | 23.14/0.6584 | 24.60/0.7267 | 25.18/0.7525 | 25.21/0.7545 | 25.44/0.7638 | 25.71/0.7784 |
| Manga109 | x2 | 30.81/0.9347 | 36.56/0.9704 | 37.16/0.9738 | 27.27/0.9731 | 37.60/0.9736 | 38.15/0.9756 |
| | x3 | 26.96/0.8559 | 31.12/0.9202 | 32.01/0.9329 | 32.19/0.9334 | 32.42/0.9359 | 32.95/0.9395 |
| | x4 | 24.91/0.7866 | 27.89/0.8590 | 28.88/0.8854 | 29.09/0.8893 | 29.18/0.8914 | 29.74/0.8965 |

**Table 2** Average PSNR/SSIM for the upscaling factors ×2, ×3 and ×4 on datasets Set5, Set14, B100, Urban100 and Manga109. All models are trained on the images from DIV2K, Flickr and ImageNet datasets.

| Dataset | Scale | Bicubic | MemNet | EDSR | D-DBPN | SRFBN | Proposed |
|---|---|---|---|---|---|---|---|
| Set5 | x2 | 33.66/0.9299 | 37.78/0.9597 | 38.11/0.9602 | 38.09/0.9600 | 38.11/0.9609 | 38.10/0.9606 |
| | x3 | 30.39/0.8677 | 34.09/0.9248 | 34.65/0.9280 | -/- | 34.70/0.9292 | 34.71/0.9287 |
| | x4 | 28.42/0.8099 | 31.74/0.8893 | 32.46/0.8968 | 32.47/0.8980 | 32.47/0.8983 | 32.40/0.8972 |
| Set14 | x2 | 30.23/0.8689 | 33.28/0.9142 | 33.92/0.9195 | 33.85/0.9190 | 33.82/0.9196 | 33.86/0.9189 |
| | x3 | 27.54/0.7742 | 30.00/0.8350 | 30.52/0.8462 | -/- | 30.51/0.8461 | 30.53/0.8464 |
| | x4 | 26.00/0.7026 | 28.26/0.7723 | 28.80/0.7876 | 28.82/0.7860 | 28.81/0.7868 | 28.79/0.7867 |
| B100 | x2 | 29.55/0.8438 | 32.08/0.8978 | 32.32/0.9013 | 32.27/0.9000 | 32.29/0.9010 | 32.27/0.9016 |
| | x3 | 27.21/0.7397 | 28.96/0.8001 | 29.25/0.8093 | -/- | 29.24/0.8084 | 29.29/0.8095 |
| | x4 | 25.96/0.6693 | 27.40/0.7281 | 27.71/0.7420 | 27.72/0.7400 | 27.72/0.7409 | 27.80/0.7417 |
| Urban 100 | x2 | 26.88/0.8406 | 31.31/0.9195 | 32.93/0.9351 | 32.55/0.9324 | 32.62/0.9328 | 32.83/0.9336 |
| | x3 | 24.46/0.7354 | 27.56/0.8376 | 28.80/0.8653 | -/- | 28.73/0.8641 | 28.82/0.8658 |
| | x4 | 23.14/0.6584 | 25.50/0.7630 | 26.64/0.8033 | 26.38/0.7946 | 26.60/0.8015 | 26.61/0.8024 |
| Manga109 | x2 | 30.81/0.9347 | 37.72/0.9740 | 39.10/0.9773 | 38.89/0.9774 | 39.08/0.9779 | 39.15/0.9786 |
| | x3 | 26.96/0.8559 | 32.51/0.9369 | 34.17/0.9467 | -/- | 34.18/0.9481 | 34.27/0.9485 |
| | x4 | 24.91/0.7866 | 29.42/0.8942 | 31.02/0.9148 | 30.91/0.9137 | 31.15/0.9160 | 31.14/0.9155 |